\documentclass{article}
\usepackage{ijcai19}

\usepackage{times}
\usepackage{soul}
\usepackage{url}
\usepackage[hidelinks]{hyperref}
\usepackage[utf8]{inputenc}
\usepackage[small]{caption}
\usepackage{graphicx}
\usepackage{amsmath}
\usepackage{booktabs}
\usepackage[boxed,vlined]{algorithm2e}
\usepackage{amsmath,amssymb,amsfonts,dsfont}  
\usepackage[usenames,dvipsnames]{color}
\usepackage{xcolor}

\usepackage{tikz}
\usetikzlibrary{shapes.geometric}
\usetikzlibrary{positioning,shapes,shadows,arrows,calc}
\usetikzlibrary{intersections}
\pgfdeclarelayer{background}
\pgfdeclarelayer{foreground}
\pgfsetlayers{background,main,foreground}

\RestyleAlgo{algoruled}
\DontPrintSemicolon
\LinesNumbered
\SetAlgoVlined
\SetFuncSty{textsc}

\SetKwInOut{Input}{Input}
\SetKwInOut{Output}{Output}
\SetKw{Return}{return}

\DeclareMathOperator*{\argmin}{arg\,min}
\DeclareMathOperator*{\argmax}{arg\,max}

\newcommand{\surro}{\hat{f}}
\newcommand{\X}{\mathcal{X}}

\title{Towards Assessing the Impact of Bayesian Optimization's Own Hyperparameters}

\author{%
\hspace{4em}Marius Lindauer\footnote{Contact Author}$^{,1}$\and Matthias Feurer$^{1}$\and Katharina Eggensperger$^{1}$\and\newline André Biedenkapp$^{1}$\And Frank Hutter$^{1,2}$\\
\affiliations
$^1$ University of Freiburg ~~~~~~~ $^2$ Bosch Center for Artificial Intelligence\\
\emails{\{lindauer, feurerm, eggenspk, biedenka, fh\}@cs.uni-freiburg.de}
}

\begin{document}

\maketitle

\begin{abstract}
Bayesian Optimization (BO) is a common approach for hyperparameter optimization (HPO) in automated machine learning. Although it is well-accepted that HPO is crucial to obtain well-performing machine learning models, tuning BO's own hyperparameters is often neglected. In this paper, we empirically study the impact of optimizing BO's own hyperparameters and the transferability of the found settings using a wide range of benchmarks, including artificial functions, HPO and HPO combined with neural architecture search. In particular, we show (i) that tuning can improve the any-time performance of different BO approaches, that optimized BO settings also perform well (ii) on similar problems and (iii) partially even on problems from other problem families, and (iv) which BO hyperparameters are most important. 

\end{abstract}

\section{Introduction}
\label{sec:intro}

Due to its sample efficiency, Bayesian Optimization (BO) is a popular approach for optimizing the hyperparameters of machine learning algorithms~\cite{snoek-nips12a,bergstra-scipy13a,thornton-kdd13a,feurer-nips2015a,jin-arxiv18a,shahriari-ieee16a,feurer-automl2019a}.
Treating the validation loss of trained machine learning models as a black box function $f$,
we can formulate the hyperparameter optimization problem as:

\begin{equation}
    x^* \in \argmin_{x \in \mathcal{X}} f(x)
\end{equation}

where $\mathcal{X}$ is space of possible configurations $x$.

Although the community is aware of the necessity of hyperparameter optimization (HPO) for machine learning algorithms,
the impact of BO's own hyperparameters is not reported in most BO papers.
On top of this, new BO approaches (and implicitly their hyperparameters) are often developed on cheap-to-evaluate artificial functions and then evaluated on real benchmarks. Although we acknowledge that this is a reasonable protocol to prevent over-engineering on the target function family (here for example HPO benchmarks of machine learning algorithms), we
believe that it is important to study whether this practice is indeed well-founded. 

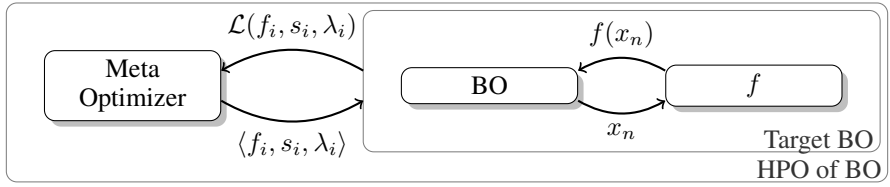
\begin{figure*}[tbh]
    \centering
    \tikzstyle{activity}=[rectangle, draw=black, rounded corners, text centered, text width=6em, fill=white, drop shadow]
\tikzstyle{data}=[rectangle, draw=black, text centered, fill=black!10, text width=4em, drop shadow]
\tikzstyle{myarrow}=[->, thick]
\begin{tikzpicture}[node distance=10em]
	
	\node (TBO) [activity] {BO};
	\node (func) [activity, right of=TBO] {$f$};
	
%	\draw[myarrow, bend right] (TBO) -- (func);
	\path ($(TBO.east)+(0,-0.2)$) edge[bend right, ->, thick] node [below] {$x_n$} ($(func.west)+(0,-0.2)$);
	\path ($(func.west)+(0,0.2)$) edge[bend right, ->, thick] node [above] {$f(x_n)$} ($(TBO.east)+(0,0.2)$);
	
	\node (MBO) [activity, left of=TBO, node distance=13.5em] {Meta Optimizer};
	
	\path ($(MBO.east)+(0,-0.2)$) edge[bend right, ->, thick] node [below] {$\langle f_i,s_i,\lambda_i \rangle$} ($(TBO.west)+(-0.5,-0.2)$);
	\path ($(TBO.west)+(-0.5,0.2)$) edge[bend right, ->, thick] node [above] {$\mathcal{L}(f_i, s_i, \lambda_i)$} ($(MBO.east)+(0,0.2)$);
	\begin{pgfonlayer}{background}

    	\path (TBO -| TBO.west)+(-0.5,1.0) node (resUL) {};
    	\path (func.east |- func.south)+(0.5,-0.6) node(resBR) {};
    	\path [rounded corners, draw=black!50] (resUL) rectangle (resBR);
		\path (TBO.east |- TBO.south)+(3.2,-0.5) node [text=black!80] {Target BO};
    	
    	\path (MBO -| MBO.west)+(-0.5,1.1) node (resUL) {};
    	\path (func.east |- func.south)+(0.6,-1.0) node(resBR) {};
    	\path [rounded corners, draw=black!50] (resUL) rectangle (resBR);
		\path (MBO.east |- MBO.south)+(7.95,-0.64) node [text=black!80] {HPO of BO};

    \end{pgfonlayer}
\end{tikzpicture}
    \caption{Using a meta-optimizer to optimize the hyperparameters of a target Bayesian Optimization system, where $f$ is the target function, $s$ and $\lambda$ are a random seed and the hyperparameters of the target-BO.}
    \label{fig:overview}
\end{figure*}

We emphasize that this paper considers HPO on 
two levels as shown in Figure~\ref{fig:overview}: (i) HPO of machine learning algorithms, which we consider as our target function (Target BO) 
and (ii) optimization of the target-BO's own choices using a meta-optimizer.

In particular, we study several research questions related to the meta-optimization problem of BO's hyperparameters:

\begin{enumerate}
    \item How large is the impact of tuning BO's own hyperparameters if one was allowed to tune these on each function independently? 
    \item How well does the performance of an optimized configuration of the target-BO generalize to similar new functions from the same family?
    \item How well does the performance of an optimized configuration of the target-BO generalize to new functions from different families?
    \item Which hyperparameters of the target-BO are actually important on the benchmarks considered and how does their best setting depend on the benchmark family being considered?
\end{enumerate}

To answer these questions, we ran extensive empirical studies on different function families, including commonly used artificial functions, hyperparameter optimization on Support Vector Machines (SVMs), and hyperparameter optimization and architecture search for (small-scale) deep neural networks. Since applying BO for HPO is already expensive, meta-optimization of BO's hyperparameters can easily become infeasible. Therefore, we surrogate HPO benchmarks~\cite{eggensperger-aaai15,eggensperger-mlj18a}, as recently widely adopted in the BO-community~\cite{falkner-icml18a,perrone-neurips18a,probst-jmlr19a}.

\section{Background: Bayesian Optimization}
\label{sec:back}

\begin{algorithm}[tbp]
\Input{Search Space $\mathcal{X}$,
black box function $f$, 
acquisition function $\alpha$,
maximal number of function evaluations $m$
}
\BlankLine
$\mathcal{D}_0$ $\leftarrow$ initial\_design($\mathcal{X}$); \\
\For{n = $1, 2, \ldots m - |D_0|$}{
  $\surro$ $\leftarrow$ fit predictive model on $\mathcal{D}_{n-1}$;\\
  select $x_{n}$ by optimizing $x_{n} \in \argmax_{x \in \mathcal{X}} \alpha(x; \mathcal{D}_{n-1}, \surro)$;\\
  Query $y_{n} := f(x_{n})$;\\
  Add observation to data $D_{n} := D_{n-1} \cup \{\langle x_{n}, y_{n} \rangle \}$;\\
}
\Return{Best seen $x$}
\caption{Bayesian Optimization (BO)}
\label{algo:bo}
\end{algorithm}

Bayesian Optimization (BO) aims to minimize the number of function evaluations required to minimize black box function.
To this end, a predictive model $\surro$ is trained on all pairs $\langle x, f(x)\rangle$ of points observed so far
to estimate the unknown function $f$.
Gaussian Processes (GPs) are a common choice to model~$\surro$,
due to their closed-form tractability and their well-calibrated uncertainty estimates.
Alternatives to GPs include random forests (RFs) ~\cite{hutter-lion11a} and artificial neural networks~\cite{snoek-icml15a,springenberg-nips2016,perrone-neurips18a};
however, these are rare since obtaining efficient out-of-the-box uncertainty estimates in DNNs is still a topic of ongoing research.

Bayesian Optimization is shown in Algorithm~\ref{algo:bo}.
It starts with using an initial design to sample the search space $\X$.
In the simplest case, this can be a single, randomly drawn point $x_0$ and its corresponding function value $f(x_0)$.
The initial design gives rise to initial observation data $D_0 := \{\langle x_i, f(x_i) \rangle \}_i$.
Given the remaining budget of function evaluations (or other budgets such as time constraints),
BO iterates the following three steps: 
(i) it fits a probabilistic model~$\surro$ (typically a GP) to the previous 
observations $D_{n-1}$, 
(ii) it optimizes an acquisition function $\alpha$, trading off exploration and exploitation, to select the next point~$x_n$, 
(iii) it queries $x_n$ to obtain $f(x_n)$ and adds it to $D_{n-1}$.
In the end, the algorithm typically returns the best seen $x$, the so-called incumbent.\footnote{We note, that there are BO variants, for example when using entropy search~\cite{hennig-jmlr12a}, that return an incumbent based on the predictive model $\hat{f}$ rather than $D_n$.}

To select the next point in each iteration,
different acquisition functions can be optimized.
A well known acquisition function is the expected improvement (EI) criterion~\cite{mockus-tgo78a}.
An advantage of EI is that it has a closed-form solution making it efficient to compute:

\begin{eqnarray}
    E[I(x)] &= \mathbb{E} [ \max \{0, f^* - f(x)\}] \nonumber\\
            &= \sigma_x \cdot [u \cdot \Phi(u) + \phi(u)] 
\end{eqnarray}

\noindent where $u := \frac{f^* - \mu_x}{\sigma_x}$, $\mu_x$ and $\sigma_x$ are the mean and variance predicted by $\surro$, $f^*$ is the best function value observed so far, and 
$\phi$ and $\Phi$ are the PDF and CDF of a normal distribution, resp. 
For an overview on BO and other acquisition functions, we refer the interested reader to recent overview papers \cite{brochu-arXiv10a,shahriari-ieee16a,frazier-arxiv18a}.

\section{Related Work}
\label{sec:rel}

Already Jones et al. \shortcite{jones-jgo98a} pointed out that the performance of global optimization algorithms can be improved by considering several design decisions; for example, they recommended to use different transformations of the response values to improve the performance of BO. Similar studies for example by Brockhoff et al. \shortcite{brockhoff-gecco15a} and Morar et al. \shortcite{morar-dso17a} evaluated the impact of the initial design, while Picheny et al. \shortcite{picheny-smo2013a} evaluated different acquisition functions for noisy benchmark problems.

Snoek et al. \shortcite{snoek-icml15a} used  HPO to tune the architecture of the DNN in their DNGO approach, but they did not report the impact of their HPO and on which functions they exactly evaluated it. Given the importance of a good DNN architecture and its hyperparameter configuration, we expect that their HPO was crucial to achieve good performance.

Dang et al. \shortcite{dang-gecco17a} performed a study similar to ours, but they studied how well surrogate benchmarks can be used for meta-tuning the algorithm configurator irace. In contrast to Dang et al, we focus here on BO and the problem of generalization of tuned BO configurations to other functions. Furthermore, we do not claim that surrogate benchmarks are a perfect replacement for real benchmarks functions, but we use them to approximate real-world problem with different surfaces to better understand the tunability of BO.

Slightly related to our study here are also recent efforts of automating design decisions of BO~\cite{hoffman-uai11a,grosse-uai12a,duvenaud-icml13,malkomes-nips16}, most of the time related to better predictive models. For example, Malkomes and Garnett \shortcite{malkomes-nips18a} proposed to use BO inside of the BO algorithm to improve the quality of the predictive model. Our work is orthogonal to theirs since they aim to locally improve the performance of a BO algorithm whereas we try to provide insights on the impact of BO's own hyperparameters.

\section{Hyperparameter Optimization for\newline Bayesian Optimization}
\label{sec:hpo}

To optimize the hyperparameters of the target-BO, we first need to define the loss metric to be optimized by the meta-optimizer.
Because hitting the optimum as accurately as possible is important in many BO applications, we consider log-regret to the actual optimum of the target function. 
Although the optimum is not known in many applications, for the sake of our empirical study we assume that the optimum is given for the functions we used for training.
The target-BO does not have access to the optimum and can only observe $f(x)$; the log-regret is only used for the meta-optimizer.

Another important characteristic of BO is a good anytime performance. 
Work on learning to learn~\cite{chen-icml17a} and learning to optimize~\cite{li-iclr17a} has shown that the performance over time yields a stronger signal than only considering the end point of an optimization trajectory.
Therefore, we consider the log-regret averaged over time. 

We also take into account that a robust BO approach should not only perform well on a single function, but at least on a family of similar functions, e.g., optimizing the hyperparameters of one algorithm on different datasets. Therefore, we are interested in optimizing the loss in expectation over a distribution of functions.

This leads us to the following meta-loss of our target-BO optimized by the meta-optimizer:

\begin{equation}
    \mathcal{L}(\lambda) = \mathbb{E}_{f\sim\mathcal{F}}\left[\frac{1}{T} \sum_{t=1}^T  \min_{\hat{x} \in \mathbf{x}(\lambda)_{1:t}}  \log \left( f(\hat{x}) - f(x^*) \right) \right]
    \label{eq:loss}
\end{equation}

where $\mathcal{F}$ is a distribution over functions $f$, $x^*$ is the optimum of $f$, $T$ is the budget of allowed function evaluations and
$\lambda$ is a hyperparameter configuration of the target-BO used to obtain a sequence of samples $\mathbf{x}_{1:t}$ (one for each time step).\footnote{In practice, we also take randomness of the BO algorithm into account, but omit averaging over multiple seeds to not clutter notation. } 

\subsection{Meta-Optimization}

Our meta-optimization problem itself is a hyperparameter optimization problem.
A straightforward approach would be to apply BO as meta-optimizer to tune the target-BO~\cite{snoek-icml15a}.
However, since we are interested in a configuration $\lambda^*$ that performs well across a set of functions,
we would need to evaluate the target-BO for each configuration $\lambda$ on all functions (if we can assume that we approximate the distribution of functions by a finite set of functions). Even if the target functions would be fairly cheap (e.g., artificial functions such as Branin), 
the meta-optimization would be very expensive because of the overhead induced by BO, i.e., predictive model training and acquisition function optimization.

To efficiently optimize our loss $\mathcal{L}(\lambda)$, we use approaches from algorithm configuration.
As recently argued by Eggensperger et al. \shortcite{eggensperger-mlj18a}, algorithm configuration is a variant of hyperparameter optimization.
Algorithm configuration aims to minimize an algorithm's cost $c(\lambda,\pi)$ w.r.t. a hyperparameter configuration $\lambda$  across a set of so-called instances~$\pi \sim \Pi$:

\begin{equation}
    \lambda^* \in \argmin_{\lambda \in \Lambda} \mathbb{E}_{\pi \sim \Pi}\left[c(\lambda, \pi)\right]
\end{equation}

By instantiating instances $\pi$ with functions $f$, we can directly apply this formulation to attack our optimization problem given in Equation~\ref{eq:loss}.

\subsection{Hyperparameters in Bayesian Optimization}
\label{sub:hpo_bo}

Bayesian Optimization has many hyperparameters and design options.
These include but are not limited to:
the initial design (e.g., a Sobol sequence~\cite{snoek-nips12a}, random points~\cite{hutter-lion11a}, or latin-hypercube design~\cite{brockhoff-gecco15a}),
the acquisition function (e.g., probability of improvement~\cite{kushner-jfe64a}, expected improvement~\cite{mockus-tgo78a} or upper confidence bounds~\cite{srninivas-icml10a}), transformations of function value observations~\cite{jones-jgo98a}, the predictive model and its hyperparameters (e.g., GPs or RFs~\cite{hutter-lion11a}), and sometimes even interleaved random sampling is used~\cite{bull-jmlr11a,hutter-lion11a,ahmed-bo16a}.

Based on these design options, we created three design spaces. To allow for a comprehensive and still manageable study, each of these spaces includes exactly one predictive model family and its hyperparameters: GP with maximum likelihood (GP-ML)~\cite{jones-jgo98a}, with maximum a posteriori (GP-MAP)~\cite{snoek-nips12a} and random forests (RF)~\cite{hutter-lion11a}. Although random forests are much less commonly used compared to GPs for BO, we included them here to study a complementary approach. Our BO-implementation and its default settings are inspired by the BO-tools Spearmint~\cite{snoek-nips12a}, RoBO~\cite{klein-bayesopt17a}, GpyOpt~\cite{gpyopt2016} and  SMAC~\cite{hutter-lion11a}. We note that all our models can handle continuous, categorical and conditional search spaces either natively (RF) or by using specialized kernels (GP)~\cite{levesque-ieee17a}. 
All three design spaces cover the initial design, the acquisition function, function value transformations and interleaved random sampling.\footnote{Although we cover quite many design options giving rise to hyperparameter spaces with $15$ to $24$ hyperparameters,
there are many more options which could be included, e.g., other predictive models~\cite{snoek-icml15a,springenberg-nips2016,perrone-neurips18a}, other acquisition functions~\cite{hennig-jmlr12a,hernandez-nips14a,wu-nips16a} or how to optimize the acquisition function itself. To have a manageable meta-optimization problem which still allows insights, we focused here on only some BO approaches.}

\section{Experiments}
\label{sec:exp}

We now discuss the results of studying the research questions presented in Section~\ref{sec:intro}.

\begin{table*}[h]
    \centering
    \begin{tabular}{lrrrrr}
         \toprule
         Family & \#Functions & \#Cont. & \#Cat. & Conditionals & Ref.\\
         \midrule
         Artificial functions & $10$ & $2-6$ & $0$ & $\times$ & --\\
         HPO SVM & $10$ & $3$ & $1$ & $\checkmark$ & \cite{kuhn-arxiv2018a}\\
         HPO+NAS ParamNet & $6$ & $6$ & $0$ & $\times$ & \cite{falkner-icml18a}\\
         \bottomrule
    \end{tabular}
   \caption{Benchmarks from HPOlib with the family name, the number of benchmarks in each family, the number of continuous hyperparameters, the number of categorical hyperparameters, whether conditional dependencies between hyperparameters exist and a reference.}
    \label{tab:families} 
\end{table*}

\subsection{Experimental Setup}

As function families, we chose three different, but fairly typical benchmarks for BO, see Table~\ref{tab:families} 
First, we used commonly used artificial functions, including Branin or Hartmann3.
These functions are continuous and have 2 to 6 dimensions.
Second, we used benchmarks from an experimental study by Kühn et al. \shortcite{kuhn-arxiv2018a}
for optimizing the hyperparameters of SVMs on 10 different datasets from OpenML.
In contrast to the artificial functions, there is one categorical hyperparameter for choosing the kernel
and two of the continuous hyperparameters are only active based on the chosen kernel.
Third, we used benchmarks for optimizing small neural networks (including their architecture size, dropout rate, batchsize and learning rate).
The latter two benchmarks are surrogate benchmarks~\cite{eggensperger-aaai15} and thus we can fairly well approximate their optima.

We ran all benchmarks on a high-performance cluster equipped with Intel Xeon E5-2630v4 processors and 128GB RAM.
We let the target-BO evaluate $100$ function evaluations to obtain a loss value for the meta-optimizer.
As a meta-optimizer we chose to use SMACv3~\cite{smac-2017}
and used a maximum runtime of 2 days to ensure that the meta-optimization can collect sufficient amounts of data.
We ran the meta-optimizer 20 times with different random seeds whereas the first 10 runs used the standard SMAC mode
and the second 10 runs used the random sampling of SMAC.
We validated the best target-BO configuration found overall\footnote{We note that we can only do that because we do not compare meta-optimization approaches, but we are solely interested in the potential peak performance of the target-BO.},
by running it again on each function 20 times with different random seeds.
In the following, we report aggregated results across all functions in a family.

\subsection{Q1: Impact of Hyperparameter Optimization for Bayesian Optimization}

\begin{figure*}[tbp]
    \centering
    \includegraphics[width=0.48\textwidth]{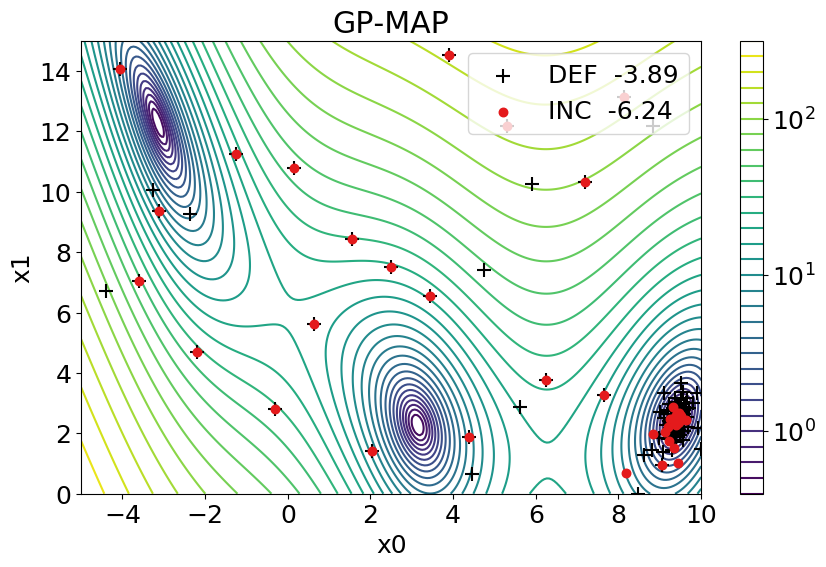}\quad%
    \includegraphics[width=0.48\textwidth]{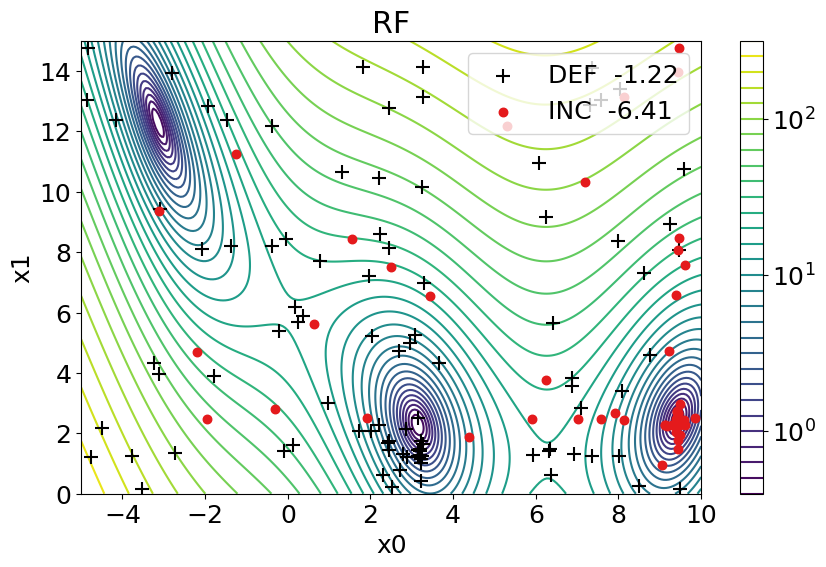}
    \caption{Sampling behavior of BO with GP-MAP (left) and RF (right) on the function Branin, including the samples of the default configurations ("DEF") and the tuned incumbent configurations ("INC"). The final log-regret is shown in the legend.}
    \label{fig:sampling}
\end{figure*}

\begin{table*}[t]
    \centering
    \begin{tabular}{l|@{\hspace*{1mm}}rrrr@{\hspace*{1mm}}|@{\hspace*{1mm}}rrrr@{\hspace*{1mm}}|@{\hspace*{1mm}}rrrr}
    {} & DEF & LOFO & ALL & IND & DEF & LOFO & ALL & IND & DEF & LOFO & ALL & IND \\
    \cmidrule{2-13}
     & \multicolumn{4}{c@{\hspace*{1mm}}|@{\hspace*{1mm}}}{RF} & \multicolumn{4}{c@{\hspace*{1mm}}|@{\hspace*{1mm}}}{GP-ML} & \multicolumn{4}{c}{GP-MAP} \\
    \midrule
    artificial & $-0.19$ & $\mathbf{-0.95}$ & $\mathbf{-1.52}$ & $\mathbf{-2.44}$ & $-2.35$ & $-2.50$ & $\mathbf{-2.80}$ & $\mathbf{-3.40}$ & $-2.41$ & $-2.43$ & $\mathbf{-2.88}$ & $\mathbf{-3.73}$\\
    SVM & $-2.65$ & $-2.73$ & $\mathbf{-2.91}$ & $\mathbf{-3.22}$ & $-2.90$ & $\mathbf{-3.11}$ & $\mathbf{-3.10}$ & $\mathbf{-3.36}$ & $-2.87$ & $-2.87$ &  $\mathbf{-3.08}$ & $\mathbf{-3.40}$ \\
    ParamNet & $-2.12$ & $\mathbf{-2.37}$ & $\mathbf{-2.39}$ & $\mathbf{-2.43}$ & $-2.15$ & $\mathbf{-2.36}$ & $\mathbf{-2.39}$ & $\mathbf{-2.47}$ & $-2.25$ & $\mathbf{-2.32}$ &  $\mathbf{-2.36}$ & $\mathbf{-2.40}$ \\
    \bottomrule
            \end{tabular}
    \caption{Log-regret averaged over time across several target functions and 20 repetitions in each benchmark family. "DEF" refers to the default setting, "LOFO" to a leave-one-function-out setting within one family, "ALL" to tuning on all functions within a family jointly, and "IND" to tuning each function independently. We note that only in the "LOFO" the optimized BO configuration has to generalize to new functions. setting We highlight results which perform significantly better than the default setting across the target functions using a paired, one-sided Wilcoxon signed rank test with a significance level of $0.05$.}
    \label{tab:res}
\end{table*}

\begin{table*}[t]
    \centering
    \begin{tabular}{l|rrr|rrr|rrr}
    & {\rotatebox[origin=l]{45}{artificial}} & {\rotatebox[origin=l]{45}{SVM}} & {\rotatebox[origin=l]{45}{ParamNet}} & {\rotatebox[origin=l]{45}{artificial}} & {\rotatebox[origin=l]{45}{SVM}} & {\rotatebox[origin=l]{45}{ParamNet}} & {\rotatebox[origin=l]{45}{artificial}} & {\rotatebox[origin=l]{45}{SVM}} & {\rotatebox[origin=l]{45}{ParamNet}}\\
    \cmidrule{2-10}
        & \multicolumn{3}{c|}{RF} & \multicolumn{3}{c|}{GP-ML} & \multicolumn{3}{c}{GP-MAP}\\
    \toprule
    artificial & $-0.95$ & $-0.79$ & $-0.91$ & $-2.50$ & $-2.65$ & $-2.10^*$ & $-2.43$ & $-2.19^*$ & $-2.40^*$ \\
    SVM & $-2.62^*$ & $-2.73$ & $-2.54^*$ & $-3.01$ & $-3.11$ & $-2.93$ & $-2.76^*$ & $-2.87$ & $-3.06$ \\
    ParamNet & $\mathbf{-2.30}$ & $\mathbf{-2.27}$ & $-2.37$ & $-2.35$ & $\mathbf{-2.32}$ & $-2.36$ & $-2.37$ & $-2.34$ & $-2.32$\\
    \bottomrule
    \end{tabular}
     \caption{Log-regret averaged over time for the out-of-family setting. The columns refer to the function family used for the meta-optimization and the rows refer to the family the best configuration was applied to. Thus, we compare the numbers in each row. The diagonal refers to the setting where we tune in a leave-one-function-out setting and evaluate on the remaining function. Results are highlighted if the performance is significantly worse than the diagonal using a paired, one-sided Wilcoxon signed rank test with a significance level of $0.05$. We marked all results with a star$^*$ if the performance is worse than the used default setting (as shown in Table~2).}
    \label{tab:oof}
\end{table*}

%\note{ML: I added the stars manually. It would be nicer to extend the script to do it automatically. Furthermore, a statistical test would be also nice to verify the stars.}

Table~\ref{tab:res} shows the results for each function family and each predictive model family.
To discuss the impact of hyperparameter optimization on BO's own hyperparameters,
we compare columns "DEF" (the default hyperparameter configuration; see Section~\ref{sub:hpo_bo}), "LOFO" (tuning jointly on all except one function from the family under consideration) , "ALL" (tuning jointly on all functions of the family under consideration) and "IND" (tuning the target-BO on each function independently).
For each combination of function family, tuning setup and model, 
the performance of the tuned target-BO ("IND") improved substantially over the default-BO ("DEF"),
showing the importance of tuning BO's own hyperparameters.
We also observe slightly different rankings comparing models
before and after tuning, which underlines the importance of tuning for fair comparisons of approaches.

Furthermore, we studied the sampling behavior of the BO approaches in detail, see Figure~\ref{fig:sampling}.
On the well known benchmark function Branin,
we can observe that all default configurations do far too much exploration
by covering a lot of the space.
However, the tuned configurations are much more greedy
and exploit one of the optima much better.
Since Branin has no local optima, this is an efficient strategy.\footnote{We note that the successful greedy optimization of Branin also shows that Branin is not well-suited for studying global optimization, although Branin is a very common benchmark function in the BO-community.}

\subsection{Q2: Generalization within Benchmark Families} 

To study generalization within a benchmark family, 
we run our meta-optimizer in a leave-one-function-out (LOFO) scheme,
i.e., tuning the target BO on $n-1$ functions and evaluating it on the remaining one.
The comparison of columns "DEF" and "LOFO" in Table~\ref{tab:res} shows that
it is indeed possible to tune BO's own hyperparameter on similar functions from the same function family.
In 5 out of 9 cases, the performance is significantly better than the default settings,
but a bit worse than tuning on all functions from the same family ("ALL") or on each function independently ("IND").
The latter also shows that tuning on each function comes with the risk of over-tuning a BO approach to a single function.

\subsection{Q3: Generalization to New Benchmark Families} 

In Table~\ref{tab:oof}, we show how well a tuned target-BO configuration performs on a different function family.
We note that all our three function families are similar to some degree.
The artificial functions and the ParamNet families are both low-dimensional, continuous functions;
however, the ParamNet functions have a much smaller range of possible function values.
Although the SVM family also has only a few dimensions,
it is the only family (we considered) that has a categorical hyperparameter
and conditional dependencies.

To our surprise, out-of-family tuning led to much better results than we expected.
Most results are better than the default performance (as shown in Table~\ref{tab:res}). However, some results related to the SVM benchmarks show some potential issues for generalizing to new benchmark families, since it is our only family with categorical hyperparameters.
Nevertheless, the results also show that the performance often drops significantly (although not always substantially) compared to optimizing on the same benchmark family.
In some few cases, tuning on a different family actually led to a slightly better performance than tuning on the original family, e.g., tuning the GP-MAP on the SVM functions and applying the tuned configuration to the ParamNet functions.
This is quite surprising because intuitively we expected that tuning on functions from the same family leads to the best results.

\subsection{Q4: Most Important Hyperparameters}

\begin{figure*}[h!]
    \centering
    \includegraphics[width=0.30\textwidth]{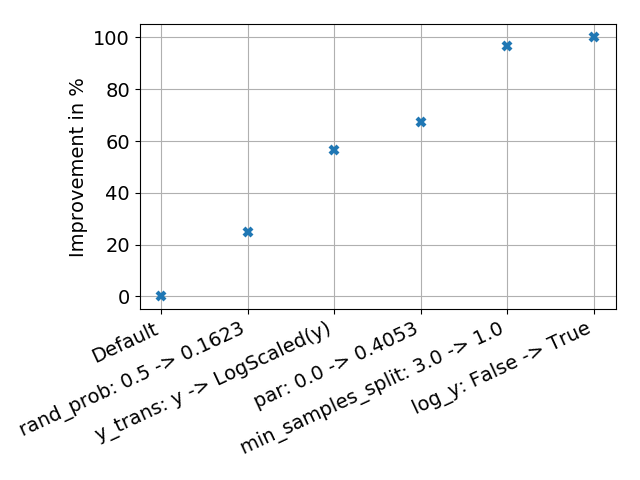}\quad
    \includegraphics[width=0.30\textwidth]{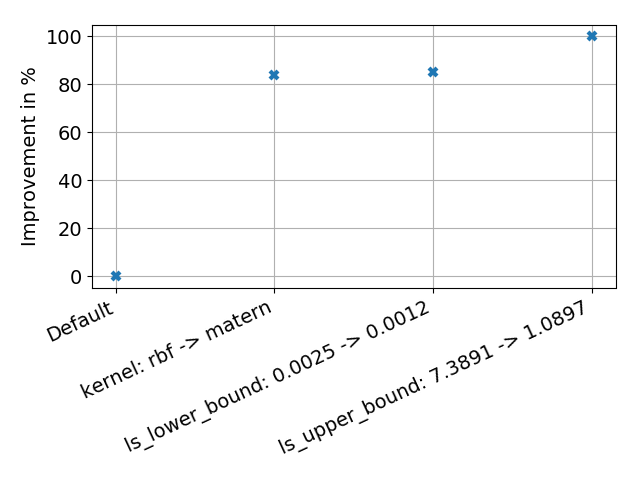}\quad
    \includegraphics[width=0.30\textwidth]{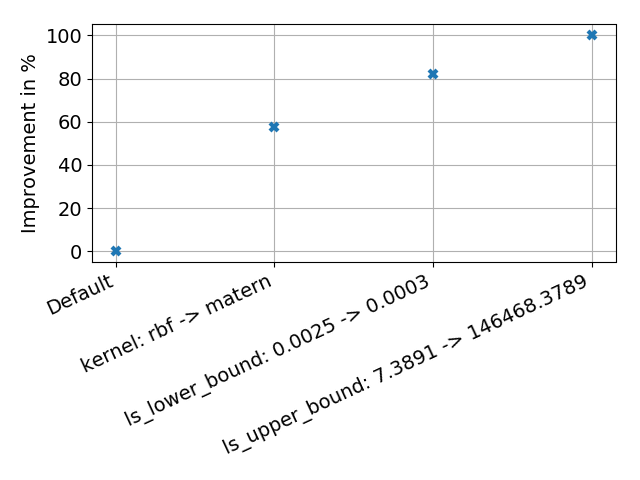}
    \caption{Results of an ablation study for RF (top), GP-ML (middle) and GP-MAP (bottom) to show which hyperparameters had to be tuned to achieve better performance on the artificial functions. An improvement of $0.0$ refers to the default performance and $1.0$ to the performance after tuning. Higher values are improvements over the configuration obtained during the tuning procedure. The x-axis shows at most $10$ hyperparameter changes which were in all cases sufficient to identify the most important hyperparameters. We show the ablation path from the default configuration to the best configuration found by tuning jointly on all functions from a family (corresponds to "ALL" in Table~\ref{tab:res}).}
    \label{fig:ablation}
\end{figure*}

To provide more insights which hyperparameters should actually be tuned, 
we ran an ablation study~\cite{fawcett-heu16a}.
Starting from the default hyperparameter configuration and moving towards the best found configuration,
this analysis iterates several times over all hyperparameters, and in each iteration, it greedily changes the most important hyperparameters from the default to its optimized setting.

Figure~\ref{fig:ablation} shows three exemplary results of this study on the artificial functions.
First, we note that not all of the tuned hyperparameters, but only a small set is important to improve performance. For the artificial functions these were always less than 6.
Second, we observed that the important hyperparameters strongly depend on the used predictive model
and also somewhat on the tuned function family.
For example, for RFs, a smaller probability of interleaved random samples (\textit{rand\_prob}; default was $50\%$, now $16\%$), a transformation of the observed function values (\textit{y\_trans}) and a shift correction of the incumbent value (\textit{par}) were important.
For GP-MAP and GP-ML, the choice of the kernel (using a Matérn kernel instead of a RBF kernel) and the bounds for the hyperpriors of the length-scale (\textit{ls\_lower\_bound} and \textit{ls\_upper\_bound}) were important.

\section{Conclusion and Future Work}
\label{sec:concl}

We believe that this work is an important first step towards hyperparameter optimization of Bayesian Optimization,
showing potential and limitations, and provides guidance which hyperparameters have to be tuned.
In all our experiments, including three different predictive models on three different function families, 
the clear conclusion was that hyperparameter optimization is important
and the default hyperparameter configurations we derived from well known tools leave ample room for improvement.

We showed that hyperparameter optimization of Bayesian Optimization is in fact possible in a leave-one-function-out manner,
indicating that hyperparameter optimization can be jointly conducted on similar functions and applied to new functions from the same family.
Even tuning on a different family of functions often worked surprisingly well, but it also comes with some risks.
We expect that it is possible to construct function families that are quite different
such that hyperparameter optimization will not generalize to new function families, e.g., the best BO setting for discrete functions with many categorical hyperparameters
will most likely not resemble those for continuous, low-dimensional functions.
In future work, we will study under which characteristics of the functions this generalization to new functions will fail.

Although our results show the importance of tuning the hyper\-parameters of Bayesian Optimization,
our empirical study is based on assumptions that do not hold in most practical applications:
(i) the optimum of the function is known and (ii) the target-function is cheap-to-evaluate.
Therefore, in future work we will develop a practical approach,
which does not require these assumptions.
Since Bayesian Optimization is an iterative any-time algorithm,
multi-fidelity optimization~\cite{falkner-icml18a} will be a promising direction.
Another idea for future work is to combine simulated function evaluations based on surrogate models with interleaved real HPO function evaluations, following a recent combination of simulated function evaluations and physical experiments~\cite{marco-icra17a}.

\paragraph*{Acknowledgments}
The authors acknowledge funding by the Robert Bosch GmbH, support by the state of Baden-Württemberg through bwHPC and the German Research Foundation (DFG) through grant no. INST 39/963-1 FUGG.

\bibliographystyle{named}
\bibliography{shortstrings,lib,local,shortproc}

\end{document}

% \nipsfinalcopy is no longer used

\maketitle

\appendix
%%%%%%%%%%%%%%%%%%%%%%%%%%%%%%%%%%%%%%%%%%%%%%%%%%%%%%%%%%%%%%%%%%%%%%%%%%
\section{Artificial Functions}
\label{appsec:arti}
%%%%%%%%%%%%%%%%%%%%%%%%%%%%%%%%%%%%%%%%%%%%%%%%%%%%%%%%%%%%%%%%%%%%%%%%%

As artificial functions, we use the functions listed in Table~\ref{tab:funcs}. 

\begin{table}[h]
    \centering
    \caption{Artificial test function in HPOlib2 with the number of dimensions, the number of local and global optima. We note that all global optima are per definition also local optima and therefore counted in both columns here.}
    \begin{tabular}{ll rrr}
    \toprule
    Function    & Short & \# Dim. & \# Local & \# Global \\
    \midrule
    Branin      & Bra & 2 & 3 & 3 \\
    Camelback   & Cam & 2 & 6 & 2\\
    GoldsteinPrice & Gold & 2 & 2 & 1 \\
    Hartmann3   & Har3  & 3 & 4 & 1\\
    Hartmann6   & Har6  & 6 &  6 & 1\\
    Levy2D      & Lev2 & 2 & $> 10$& 1\\
    Rosenbrock2D  & Ros2 & 2 & 1 & 1 \\
    Rosenbrock5D  & Ros5 & 5 & 1 & 1 \\
    SinOne      & Sin1  & 1 & 7 & 1\\
    SinTwo      & Sin2  & 2 & $> 10$ & 10\\
    \bottomrule
    \end{tabular}
    \label{tab:funcs}
\end{table}

%%%%%%%%%%%%%%%%%%%%%%%%%%%%%%%%%%%%%%%%%%%%%%%%%%%%%%%%%%%%%%%%%%%%%%%%%%
\section{OpenML datasets}
\label{appsec:openml}
%%%%%%%%%%%%%%%%%%%%%%%%%%%%%%%%%%%%%%%%%%%%%%%%%%%%%%%%%%%%%%%%%%%%%%%%%%

Among others \cite{kuhn-arxiv2018a} did a empirical study on the OpenML datasets as shown in Table~\ref{tab:openML}, which we used for tuning the hyperparameters of an SVM.

\begin{table}[h]
    \centering
    \caption{OpenML datasets for HPO SVM benchmarks}
    \begin{tabular}{llrr}
        \toprule
         ID & Name & \#Samples & \#Features  \\
         \midrule
         3 & kr-vs-kp & 3\,196 & 27 \\
         44 & spambase & 4\,601 & 58 \\
         151 & electricity & 45\,312 & 9 \\
         1038 & sylva\_agnostic & 14\,395 & 217 \\
         1043 & ada\_agnostic & 4\,562 & 49 \\
         1046 & mozilla4 & 15\,545 & 6 \\
         1050 & pc & 1\,563 & 38\\
         1120 & MagicTelescope & 19\,020 & 12 \\
         1461 & bank-marketing & 45\,211 & 17\\
         \bottomrule
    \end{tabular}
    \label{tab:openML}
\end{table}

\begin{table}[h]
\centering
\caption{HPO SVM benchmark's configuration Space with 2 conditionals}
\begin{tabular}{lrrrr}
\toprule
Name & Type & Domain & Default & Parents  \\
\midrule
cost & uniform float & $[0.0009, 1024.0]$ & 1.0 & $\times$\\ 
kernel & categorical & $\{$linear,polynomial,radial$\}$ & linear & $\times$\\ 
degree & uniform int & $[2, 5]$ & 4 & $$\checkmark$$\\ 
gamma & uniform float & $[0.0009, 1024.0]$ & 1.0 & $$\checkmark$$\\
\bottomrule
\end{tabular}
\label{tab:cs}
\end{table}

%%%%%%%%%%%%%%%%%%%%%%%%%%%%%%%%%%%%%%%%%%%%%%%%%%%%%%%%%%%%%%%%%%%%%%%%%%
\section{ParamNet Benchmarks}
\label{appsec:parament}
%%%%%%%%%%%%%%%%%%%%%%%%%%%%%%%%%%%%%%%%%%%%%%%%%%%%%%%%%%%%%%%%%%%%%%%%%%

For tuning the accuracy of deep neural networks, we have used the so-called ParamNet benchmarks from HPOlib. The DNN was optimized on the following different datasets:

\begin{itemize}
    \item Adult
    \item Higgs
    \item Letter
    \item MNIST
    \item OptDigits
    \item Poker
\end{itemize}

\begin{table}[h]
\centering
\begin{tabular}{lrrrr}
\toprule
Name & Type & Domain & Default & Parents  \\
\midrule
initial log learning rate & uniform float & $[-6.0, -2.0]$ & -4.0 & $\times$\\ 
log batch size & uniform float & $[3.0, 8.0]$ & 5.5 & $\times$\\ 
log average number of neurons & uniform float & $[4.0, 8.0]$ & 6.0 & $\times$\\ 
log final learning rate & uniform float & $[-4.0, 0.0]$ & -2.0 & $\times$\\ 
number of layers & uniform float & $[1.0, 5.0]$ & 3.0 & $\times$\\ 
dropout rate & uniform float & $[0.0, 0.5]$ & 0.25 & $\times$\\
\bottomrule
\end{tabular}
\caption{ParamNet's Configuration Space. The column \textit{Parents} indicates hyperparameters that are conditional on one other hyperparameter and are therefore not always active.}
\label{tab:cs}
\end{table}

%%%%%%%%%%%%%%%%%%%%%%%%%%%%%%%%%%%%%%%%%%%%%%%%%%%%%%%%%%%%%%%%%%%%%%%%%%
\section{Configuration Spaces for Variations of Bayesian Optimization}
\label{appsec:psc}
%%%%%%%%%%%%%%%%%%%%%%%%%%%%%%%%%%%%%%%%%%%%%%%%%%%%%%%%%%%%%%%%%%%%%%%%%%

In the following we briefly describe the configuration spaces we tuned for each BO variant and which default setting we used to compare against. Each of our four configuration spaces consists of nine hyperparameters shown in Table~\ref{tab:cs:general} which are independent of the model plus up to $16$ hyperparameters configuring the model used internally within BO.

\begin{table}[h]
\centering
\caption{General hyperparameters for Bayesian Optimization. The column \textit{Parents} indicates hyperparameters that are conditional on one other hyperparameter and are therefore not always active. The range and default value for the parameter \textit{rand\_prob} is different depending on whether RFs or GPs are used as an internal model.}
\begin{tabular}{lrrrr}
\toprule
Name & Type & Domain & Default & Parents  \\
\midrule
acq\_func & categorical & \{LCB,EI,PI\} & EI & $\times$\\ 
init\_design & categorical & $\{$LHD,Sobol,Random$\}$ & Random & $\times$\\ 
n\_configs\_x\_params & uniform int & $[1, 10]$ & 1 & $\times$\\ 
y\_trans & categorical & $\{$InvScaled(y),LogScaled(y),y$\}$ & y & $\times$\\ 
lcb\_par & uniform float & $[0.01, 100.0]$ & 0.1 & $$\checkmark$$\\ 
par & uniform float & $[0.0, 1.0]$ & 0.0 & $$\checkmark$$\\ 
scale\_inv\_perc & uniform int & $[1, 50]$ & 5 & $$\checkmark$$\\ 
scale\_log\_perc & uniform int & $[1, 50]$ & 5 & $$\checkmark$$\\
\midrule
(RF only) rand\_prob & uniform float & $[0.0, 0.5]$ & 0.5 & $\times$\\
(GP only) rand\_prob & uniform float & $[0.0, 0.1]$ & 0.0 & $\times$\\ 
\bottomrule
\end{tabular}
\label{tab:cs:general}
\end{table}

We set the default settings to resemble the default behaviour of vanilla BO using \textit{Expected Improvement} as an acquisition function, no transformations of the objective function and starting with a random initial design with $d$ samples, where $d$ is the number of dimensions of the target function. We designed our default settings based on several practical implementations and recent research on BO, but implemented all variants in the same codebase to obtain comparable and unbiased results.

Our basic configuration space allows to change to acquisition function and its hyperparameters, to apply a log or an inverse scaling~\citep{jones-jgo98a} to the objective values and to modify the type and size of the initial design. For all instantiations of a target-BO run we use iterated local search to optimize the acquisition function~\citep{levesque-ieee17a}.

%======================================================
\subsection{Bayesian Optimization using Random Forests}
%======================================================

For the BO variant using RFs we use the default setting of \texttt{SMACv3}~\citep{smac-2017}, a well studied BO package using RFs. We allow to tune the most important hyperparameters of RFs including whether to use \textit{bootstrapping} and the \textit{number of trees} as we found these to influence the quality of the model and hence the optimization procedure.

\begin{table}[h]
\centering
\caption{General hyperparameters for \textbf{Bayesian Optimization using Random Forests}.}
\begin{tabular}{lrrrr}
\toprule
Name & Type & Domain & Default & Parents  \\
\midrule
do\_bootstrapping & categorical & $\{$True,False$\}$ & True & $\times$\\ 
min\_samples\_leaf & uniform int & $[1, 100]$ & 3 & $\times$\\ 
min\_samples\_split & uniform int & $[1, 100]$ & 3 & $\times$\\ 
num\_trees & uniform int & $[2, 100]$ & 10 & $\times$\\ 
log\_y\_in\_tree & categorical & $\{$True,False$\}$ & False & $\times$\\ 
ratio\_features & uniform float & $[0.5, 1.0]$ & 0.8333333333 & $\times$\\ 
\bottomrule
\end{tabular}
\label{tab:cs:rf}
\end{table}

%======================================================
\subsection{Bayesian Optimization using Gaussian Processes}
%======================================================

The BO variants using GPs are based on the implementation of \texttt{Spearmint}\footnote{\url{https://github.com/HIPS/Spearmint}}~\citep{snoek-nips12a}, \texttt{RoBO}\footnote{\url{https://github.com/automl/RoBO/}}~\citep{klein-bayesopt17} and \texttt{GPyOpt}\footnote{\url{https://github.com/SheffieldML/GPyOpt}}~\citep{gpyopt2016}. We compare maximum likelihood (GP-ML) and maximum a posteriori (GP-MAP) point estimates of the hyperparameters to using MCMC to integrating out the hyperparameters of the GP (GP-MCMC). 

We use a simple radial basis function kernel as a default kernel, which can be replaced by a Matérn kernel. For all kernels we handle categorical inputs using the Hamming distance~\citep{hutter-phd} and conditional inputs by disabling the covariance for subspaces that do not match~\citep{levesque-ieee17a}. Furthermore, we can turn on automatic relevance detection to handle non-isotropic lengthscales and enable noise estimation to adapt to non-deterministic benchmark problems.

To find good hyperparameters for the GP models, we use lognormal and horseshoe hyperpriors for the covariance amplitude (\textit{*\_cov\_ampl\_*}), the lengthscale (\textit{*\_ls\_*}) and noise (\textit{*\_noise\_*}) as in \texttt{Spearmint}, but additionally also add the more general Gamma prior. Additionally, we allow to tune the lower and upper bounds for the hyperpriors, where the default values stem from~\texttt{RoBO}. We defined reasonable ranges for these based on initial experiments. For GP-MAP and GP-ML we use \textit{L-BFGS-B}~\citep{zhu-acm97a} using gradients to find well performing hyperparameters.

\begin{table}[h]
\centering
\caption{General hyperparameters for \textbf{Bayesian optimization using GP-ML} (upper part)\textbf{, GP-MAP and GP-MCMC} (upper and lower part). The column \textit{Parents} indicates hyperparameters that are conditional on another hyperparameter and are therefore not always active.}\label{tab:cs:gp-map}
\begin{tabular}{lrrrr}
\toprule
Name & Type & Domain & Default & Parents  \\
\midrule
ard & categorical & $\{$True,False$\}$ & False & $\times$\\ 
kernel & categorical & $\{$matern,rbf$\}$ & rbf & $\times$\\ 
ls\_lower\_bound & uniform float & $[6.1e^{-6}, 0.999]$ & 0.002 & $\times$\\ 
ls\_upper\_bound & uniform float & $[1.0, 162754.79]$ & 7.389 & $\times$\\ 
tune\_noise & categorical & $\{$True,False$\}$ & False & $\times$\\ 
\midrule
prior\_cov\_ampl & categorical & $\{$Lognormal,Gamma$\}$ & Lognormal & $\times$\\ 
prior\_ls & categorical & $\{$None,Gamma$\}$ & None & $\times$\\ 
prior\_noise & categorical & $\{$Horseshoe,Gamma$\}$ & Horseshoe & $\times$\\ 
prior\_cov\_ampl\_gamma\_a & uniform float & $[0.1, 5.0]$ & 2.0 & $$\checkmark$$\\
prior\_cov\_ampl\_gamma\_scale & uniform float & $[0.01, 1.0]$ & 0.1 & $$\checkmark$$\\ 
prior\_cov\_ampl\_lognormal\_sigma & uniform float & $[0.01, 2.0]$ & 1.0 & $$\checkmark$$\\ 
prior\_ls\_gamma\_a & uniform float & $[0.1, 5.0]$ & 2.0 & $$\checkmark$$\\ 
prior\_ls\_gamma\_scale & uniform float & $[0.01, 1.0]$ & 0.1 & $$\checkmark$$\\ 
prior\_noise\_gamma\_a & uniform float & $[0.1, 5.0]$ & 2.0 & $$\checkmark$$\\ 
prior\_noise\_gamma\_scale & uniform float & $[0.01, 1.0]$ & 0.1 & $$\checkmark$$\\ 
prior\_noise\_horseshoe\_scale & uniform float & $[0.01, 2.0]$ & 0.1 & $$\checkmark$$\\ 
\bottomrule
\end{tabular}
\end{table}

\clearpage
\section{Detailed Meta-Optimization Results}

\begin{table}[h]
    \centering
    \caption{Average log-regret over time for each function. The stdev is computed over 20 random seeds. "DEF" refers to the default setting, "IND" to tuning each function independently and "LOFO" to leave-one-function out.}
    \begin{tabular}{l|rrr}
    \toprule
     & \multicolumn{3}{c}{RF}\\
    Function  & DEF & LOFO & IND\\
    \midrule
    \midrule
    Branin & $-0.49 \pm 0.51$ & $-3.19 \pm 1.05$ & $-3.64 \pm 0.96$\\
    Camelback & $-0.84 \pm 0.44$ & $-3.99 \pm 0.73$ & $-4.20 \pm 0.87$\\
    GoldsteinPrice & $1.29 \pm 0.57$ & $-0.04 \pm 1.63$ & $-0.86 \pm 1.78$\\
    Hartmann3 & $-0.99 \pm 0.45$ & $-2.69 \pm 0.95$ & $-3.41 \pm 0.68$\\
    Hartmann6 & $-0.18 \pm 0.30$ & $-0.41 \pm 0.30$ & $-0.66 \pm 0.35$\\
    Levy2D & $-0.57 \pm 0.45$ & $-2.83 \pm 1.60$ & $-2.95 \pm 1.39$\\
    Rosenbrock2D & $0.89 \pm 0.64$ & $-1.02 \pm 0.53$ & $-1.26 \pm 0.51$\\
    Rosenbrock5D & $3.68 \pm 0.28$ & $2.61 \pm 0.47$ & $1.87 \pm 0.42$\\
    SinOne & $-2.50 \pm 1.12$ & $-5.26 \pm 0.16$ & $-5.40 \pm 0.08$\\
    SinTwo & $-2.15 \pm 0.50$ & $-3.03 \pm 0.87$ & $-3.85 \pm 1.06$\\
    \midrule
    SVM 3 & $-3.90 \pm 0.30$ & $-4.07 \pm 0.21$ & $-4.16 \pm 0.13$\\
    SVM 44 & $-2.16 \pm 0.34$ & $-3.35 \pm 0.13$ & $-3.40 \pm 0.11$\\
    SVM 151 & $-1.95 \pm 0.42$ & $-2.20 \pm 0.26$ & $-2.30 \pm 0.00$\\
    SVM 312 & $-2.66 \pm 0.62$ & $-3.33 \pm 0.48$ & $-3.44 \pm 0.60$\\
    SVM 1036 & $-3.90 \pm 0.50$ & $-4.48 \pm 0.07$ & $-4.57 \pm 0.18$\\
    SVM 1043 & $-2.99 \pm 0.66$ & $-3.61 \pm 0.44$ & $-3.89 \pm 0.52$\\
    SVM 1046 & $-2.29 \pm 0.49$ & $-2.58 \pm 0.73$ & $-3.06 \pm 0.39$\\
    SVM 1050 & $-1.97 \pm 0.51$ & $-2.12 \pm 0.56$ & $-2.12 \pm 0.56$\\
    SVM 1120 & $-2.20 \pm 0.49$ & $-2.32 \pm 0.31$ & $-2.55 \pm 0.74$\\
    SVM 1461 & $-2.50 \pm 0.88$ & $-2.72 \pm 0.91$ & $-2.68 \pm 0.89$\\
    \midrule
    ParamNetAdult & $-2.10 \pm 0.03$ & $-2.18 \pm 0.06$ & $-2.20 \pm 0.06$\\
    ParamNetHiggs & $-1.94 \pm 0.04$ & $-2.01 \pm 0.06$ & $-2.02 \pm 0.05$\\
    ParamNetLetter & $-1.91 \pm 0.14$ & $-2.14 \pm 0.07$ & $-2.15 \pm 0.10$\\
    ParamNetMnist & $-2.57 \pm 0.09$ & $-2.79 \pm 0.10$ & $-2.79 \pm 0.11$\\
    ParamNetOptdigits & $-1.97 \pm 0.02$ & $-2.06 \pm 0.06$ & $-2.07 \pm 0.05$\\
    ParamNetPoker & $-2.82 \pm 0.15$ & $-3.16 \pm 0.18$ & $-3.23 \pm 0.17$\\
    \bottomrule
    \end{tabular}
    \label{tab:my_label}
\end{table}

\begin{table}[h]
    \centering
     \caption{Average log-regret over time for each function. The stdev is computed over 20 random seeds. "DEF" refers to the default setting, "IND" to tuning each function independently and "LOFO" to leave-one-function out.}
    \begin{tabular}{l|rrr}
    \toprule
     & \multicolumn{3}{c}{GP ML}\\
    Function  & DEF & LOFO & IND\\
    \midrule
    \midrule
    Branin & $-3.45 \pm 0.57$ & $-4.20 \pm 0.39$ & $-4.35 \pm 0.22$\\
    Camelback & $-3.97 \pm 0.40$ & $-4.67 \pm 0.16$ & $-4.76 \pm 0.15$\\
    GoldsteinPrice & $-1.52 \pm 0.49$ & $-2.31 \pm 0.42$ & $-2.52 \pm 0.45$\\
    Hartmann3 & $-3.46 \pm 0.24$ & $-4.43 \pm 0.27$ & $-4.47 \pm 0.19$\\
    Hartmann6 & $-2.45 \pm 0.26$ & $-2.56 \pm 0.24$ & $-2.60 \pm 0.32$\\
    Levy2D & $-2.06 \pm 0.57$ & $-3.65 \pm 1.08$ & $-4.27 \pm 0.31$\\
    Rosenbrock2D & $-0.49 \pm 0.53$ & $-0.92 \pm 0.32$ & $-1.89 \pm 0.20$\\
    Rosenbrock5D & $1.47 \pm 0.19$ & $1.07 \pm 0.18$ & $0.99 \pm 0.12$\\
    SinOne & $-4.99 \pm 0.17$ & $-5.23 \pm 0.08$ & $-5.34 \pm 0.17$\\
    SinTwo & $-2.59 \pm 0.31$ & $-2.94 \pm 0.76$ & $-4.74 \pm 0.26$\\
    \midrule
    SVM 3 & $-3.32 \pm 0.11$ & $-3.53 \pm 0.07$ & $-3.67 \pm 0.03$\\
    SVM 44 & $-3.40 \pm 0.07$ & $-3.42 \pm 0.06$ & $-3.46 \pm 0.04$\\
    SVM 151 & $-2.05 \pm 0.00$ & $-2.47 \pm 0.02$ & $-2.48 \pm 0.02$\\
    SVM 312 & $-2.79 \pm 0.36$ & $-3.23 \pm 0.15$ & $-3.68 \pm 0.05$\\
    SVM 1036 & $-4.46 \pm 0.03$ & $-4.59 \pm 0.04$ & $-4.62 \pm 0.00$\\
    SVM 1043 & $-3.62 \pm 0.07$ & $-3.67 \pm 0.00$ & $-3.67 \pm 0.00$\\
    SVM 1046 & $-2.74 \pm 0.03$ & $-2.77 \pm 0.11$ & $-3.08 \pm 0.02$\\
    SVM 1050 & $-1.56 \pm 0.21$ & $-1.82 \pm 0.13$ & $-2.29 \pm 0.00$\\
    SVM 1120 & $-1.99 \pm 0.21$ & $-2.55 \pm 0.21$ & $-2.73 \pm 0.07$\\
    SVM 1461 & $-3.05 \pm 0.76$ & $-3.58 \pm 0.12$ & $-3.88 \pm 0.00$\\
    \midrule
    ParamNetAdult & $-2.04 \pm 0.00$ & $-2.15 \pm 0.06$ & $-2.22 \pm 0.04$\\
    ParamNetHiggs & $-1.87 \pm 0.01$ & $-2.00 \pm 0.03$ & $-2.02 \pm 0.03$\\
    ParamNetLetter & $-1.97 \pm 0.04$ & $-2.20 \pm 0.02$ & $-2.22 \pm 0.00$\\
    ParamNetMnist & $-2.51 \pm 0.07$ & $-2.94 \pm 0.05$ & $-2.97 \pm 0.06$\\
    ParamNetOptdigits & $-2.01 \pm 0.06$ & $-2.08 \pm 0.03$ & $-2.09 \pm 0.02$\\
    ParamNetPoker & $-2.34 \pm 0.10$ & $-2.99 \pm 0.17$ & $-3.07 \pm 0.10$\\
    \bottomrule
    \end{tabular}
    \label{tab:my_label}
\end{table}

\begin{table}[h]
    \centering
     \caption{Average log-regret over time for each function. The stdev is computed over 20 random seeds. "DEF" refers to the default setting, "IND" to tuning each function independently and "LOFO" to leave-one-function out.}
    \begin{tabular}{l|rrr}
     \toprule
     & \multicolumn{3}{c}{GP MAP}\\
    Function  & DEF & LOFO & IND\\
    \midrule
    \midrule
    Branin & $-3.71 \pm 0.42$ & $-4.46 \pm 0.32$ & $-4.69 \pm 0.14$\\
    Camelback & $-4.21 \pm 0.38$ & $-4.81 \pm 0.11$ & $-4.92 \pm 0.05$\\
    GoldsteinPrice & $-1.55 \pm 0.27$ & $-2.49 \pm 0.46$ & $-3.19 \pm 0.57$\\
    Hartmann3 & $-3.21 \pm 0.32$ & $-4.61 \pm 0.24$ & $-4.96 \pm 0.12$\\
    Hartmann6 & $-2.30 \pm 0.30$ & $-2.55 \pm 0.37$ & $-2.67 \pm 0.26$\\
    Levy2D & $-1.94 \pm 0.64$ & $-2.73 \pm 0.94$ & $-4.63 \pm 0.23$\\
    Rosenbrock2D & $-0.75 \pm 0.72$ & $-1.34 \pm 0.51$ & $-2.76 \pm 0.51$\\
    Rosenbrock5D & $1.19 \pm 0.12$ & $0.98 \pm 0.19$ & $0.69 \pm 0.07$\\
    SinOne & $-5.00 \pm 0.17$ & $-5.19 \pm 0.02$ & $-5.41 \pm 0.05$\\
    SinTwo & $-2.63 \pm 0.42$ & $-3.57 \pm 1.28$ & $-4.78 \pm 0.16$\\
    \midrule
    SVM 3 & $-3.25 \pm 0.12$ & $-3.53 \pm 0.08$ & $-3.64 \pm 0.03$\\
    SVM 44 & $-3.44 \pm 0.06$ & $-3.45 \pm 0.08$ & $-3.57 \pm 0.02$\\
    SVM 151 & $-2.06 \pm 0.00$ & $-2.48 \pm 0.02$ & $-2.52 \pm 0.01$\\
    SVM 312 & $-2.73 \pm 0.42$ & $-3.22 \pm 0.06$ & $-3.73 \pm 0.02$\\
    SVM 1036 & $-4.46 \pm 0.09$ & $-4.60 \pm 0.00$ & $-4.62 \pm 0.00$\\
    SVM 1043 & $-3.63 \pm 0.06$ & $-3.67 \pm 0.00$ & $-3.72 \pm 0.00$\\
    SVM 1046 & $-2.75 \pm 0.01$ & $-2.82 \pm 0.06$ & $-3.03 \pm 0.00$\\
    SVM 1050 & $-1.61 \pm 0.19$ & $-2.16 \pm 0.07$ & $-2.35 \pm 0.00$\\
    SVM 1120 & $-2.06 \pm 0.14$ & $-2.50 \pm 0.15$ & $-2.82 \pm 0.01$\\
    SVM 1461 & $-2.71 \pm 0.09$ & $-3.70 \pm 0.01$ & $-3.95 \pm 0.00$\\
    \midrule
    ParamNetAdult & $-2.04 \pm 0.00$ & $-2.16 \pm 0.07$ & $-2.26 \pm 0.01$\\
    ParamNetHiggs & $-1.87 \pm 0.01$ & $-1.99 \pm 0.04$ & $-2.04 \pm 0.02$\\
    ParamNetLetter & $-1.95 \pm 0.06$ & $-2.21 \pm 0.01$ & $-2.22 \pm 0.00$\\
    ParamNetMnist & $-2.64 \pm 0.11$ & $-2.95 \pm 0.05$ & $-3.02 \pm 0.00$\\
    ParamNetOptdigits & $-2.01 \pm 0.04$ & $-2.07 \pm 0.02$ & $-2.14 \pm 0.01$\\
    ParamNetPoker & $-2.40 \pm 0.10$ & $-3.02 \pm 0.12$ & $-3.13 \pm 0.05$\\
    \bottomrule
    \toprule
    \end{tabular}
    \label{tab:my_label}
\end{table}

\begin{table}[h]
    \centering
    \caption{Average log-regret over time for each function. The stdev is computed over 20 random seeds. "DEF" refers to the default setting, "IND" to tuning each function independently and "LOFO" to leave-one-function out.}
    \begin{tabular}{l|rrr}
    \toprule
     & \multicolumn{3}{c}{GP MCMC}\\
    Function  & DEF & LOFO & IND\\
    \midrule
    \midrule
    Branin & $-1.73 \pm 0.32$ & $-4.59 \pm 0.20$ & $-4.79 \pm 0.13$\\
    Camelback & $-2.39 \pm 0.40$ & $-4.67 \pm 0.24$ & $-4.88 \pm 0.16$\\
    GoldsteinPrice & $-0.52 \pm 0.40$ & $-3.01 \pm 1.42$ & $-3.26 \pm 1.06$\\
    Hartmann3 & $-2.65 \pm 0.29$ & $-4.57 \pm 0.26$ & $-4.87 \pm 0.08$\\
    Hartmann6 & $-1.77 \pm 0.17$ & $-2.30 \pm 0.50$ & $-2.81 \pm 0.43$\\
    Levy2D & $-1.56 \pm 0.29$ & $-3.20 \pm 0.47$ & $-4.18 \pm 1.05$\\
    Rosenbrock2D & $0.31 \pm 0.24$ & $-1.12 \pm 0.56$ & $-1.37 \pm 0.32$\\
    Rosenbrock5D & $2.06 \pm 0.32$ & $0.64 \pm 0.26$ & $0.57 \pm 0.22$\\
    SinOne & $-4.87 \pm 0.13$ & $-5.28 \pm 0.03$ & $-5.33 \pm 0.04$\\
    SinTwo & $-3.23 \pm 0.66$ & $-4.43 \pm 0.50$ & $-4.77 \pm 0.12$\\
    \midrule
    SVM 3 & $-3.58 \pm 0.06$ & $-3.58 \pm 0.04$ & $-3.64 \pm 0.05$\\
    SVM 44 & $-3.32 \pm 0.07$ & $-3.41 \pm 0.06$ & $-3.51 \pm 0.05$\\
    SVM 151 & $-2.11 \pm 0.21$ & $-2.25 \pm 0.19$ & $-2.33 \pm 0.13$\\
    SVM 312 & $-2.24 \pm 0.36$ & $-3.12 \pm 0.17$ & $-3.32 \pm 0.21$\\
    SVM 1036 & $-4.29 \pm 0.14$ & $-4.51 \pm 0.05$ & $-4.55 \pm 0.06$\\
    SVM 1043 & $-3.45 \pm 0.19$ & $-3.64 \pm 0.06$ & $-3.68 \pm 0.01$\\
    SVM 1046 & $-2.64 \pm 0.10$ & $-2.80 \pm 0.12$ & $-2.89 \pm 0.07$\\
    SVM 1050 & $-1.76 \pm 0.18$ & $-2.19 \pm 0.06$ & $-2.22 \pm 0.05$\\
    SVM 1120 & $-2.09 \pm 0.20$ & $-2.52 \pm 0.18$ & $-2.71 \pm 0.04$\\
    SVM 1461 & $-2.85 \pm 0.33$ & $-3.26 \pm 0.42$ & $-3.19 \pm 0.84$\\
    \midrule
    ParamNetAdult & $-2.04 \pm 0.00$ & $-2.14 \pm 0.04$ & $-2.18 \pm 0.05$\\
    ParamNetHiggs & $-1.93 \pm 0.05$ & $-2.00 \pm 0.04$ & $-2.00 \pm 0.02$\\
    ParamNetLetter & $-2.04 \pm 0.05$ & $-2.18 \pm 0.03$ & $-2.18 \pm 0.03$\\
    ParamNetMnist & $-2.72 \pm 0.10$ & $-2.92 \pm 0.06$ & $-2.92 \pm 0.06$\\
    ParamNetOptdigits & $-2.02 \pm 0.03$ & $-2.06 \pm 0.04$ & $-2.06 \pm 0.04$\\
    ParamNetPoker & $-2.74 \pm 0.18$ & $-3.02 \pm 0.09$ & $-3.06 \pm 0.10$\\
    \bottomrule
    \end{tabular}
    \label{tab:my_label}
\end{table}

\clearpage
\section{Final Log-Regret Results}

\begin{table}[h]
    \centering
    \caption{\emph{Final log-regret} across several benchmark functions and 20 repetitions in each benchmark family. "DEF" refers to the default setting,
    "LOFO" to leave-one-function-out within one family, and "IND" to tuning each function independently. We highlight results which perform significantly better than the default setting across the benchmark functions using a paired, one-sided Wilcoxon signed rank test with significance level $0.05$.}
    \begin{tabular}{l|rrr|rrr}
    \toprule
     & \multicolumn{3}{c|}{RF}& \multicolumn{3}{c}{GP ML}\\
    Family & DEF & LOFO & IND & DEF & LOFO & IND\\
    \midrule
artificial & $-0.50$ & $\mathbf{-2.73}$ & $\mathbf{-3.19}$ & $-3.16$ & $\mathbf{-3.92}$ & $\mathbf{-4.35}$\\
HPO SVM & $-3.61$ & $\mathbf{-4.21}$ & $\mathbf{-4.17}$ & $-4.34$ & $\mathbf{-4.55}$ & $\mathbf{-4.68}$\\
ParamNet & $-2.30$ & $\mathbf{-2.45}$ & $\mathbf{-2.48}$ & $-2.15$ & $\mathbf{-2.47}$ & $\mathbf{-2.50}$\\

    \bottomrule
    \toprule
    & \multicolumn{3}{c|}{GP MAP}& \multicolumn{3}{c}{GP MCMC}\\
    Family & DEF & LOFO & IND & DEF & LOFO & IND\\
    \midrule
artificial & $-3.24$ & $\mathbf{-4.07}$ & $\mathbf{-4.67}$ & $-2.24$ & $\mathbf{-4.37}$ & $\mathbf{-4.64}$\\

HPO SVM & $-4.23$ & $\mathbf{-4.64}$ & $\mathbf{-4.70}$ & $-4.23$ & $\mathbf{-4.50}$ & $\mathbf{-4.60}$\\

ParamNet & $-2.19$ & $\mathbf{-2.47}$ & $\mathbf{-2.54}$ & $-2.30$ & $\mathbf{-2.46}$ & $\mathbf{-2.47}$\\

    \bottomrule
            \end{tabular}
    \label{tab:my_label}
\end{table}

\clearpage
\section{Found Hyperparameters for each Target-BO}

\begin{table}[h]
    \centering
        \caption{Best found configuration for target-BO using RFs}
    \begin{tabular}{lr r r r}
\toprule
 & {Default}  & artificial & HPO SVM & ParamNet \\
\midrule
acq\_func & EI & PI & PI & EI\\
init\_design & Random & Sobol & LHD & LHD\\
n\_configs\_x\_params & 1 & 6 & 6 & 3\\   
y\_trans & y & InvScaled(y) & y & LogScaled(y)\\ 
lcb\_par & - & - & - & - \\ 
par & 0.0 & 0.90 & 0.57 & 0.46 \\
scale\_inv\_perc & - & 13 & - & -\\
scale\_log\_perc & - & - & - & 1\\
rand\_prob & 0.5 & 0.08 & 0.50 & 0.01 \\ 
\midrule
do\_bootstrapping & True & True & False & False\\
log\_y\_in\_tree & False & False & True & True\\
min\_samples\_leaf & 3 & 1 & 1 & 1\\
min\_samples\_split & 3 & 4 & 61 & 1\\        
num\_trees & 10 & 3 & 40 & 7\\
ratio\_features & 0.83 & 0.90 & 0.70 & 0.73 \\       
\bottomrule
    \end{tabular}
    \label{tab:my_label}
\end{table}

\begin{table}[h]
    \centering
    \caption{Best found configuration for target-BO using GP-ML. Values were rounded to the second or fourth digit for increased readability.}
    \begin{tabular}{lr r r r}
\toprule
 & {Default}  & artificial & HPO SVM & ParamNet \\
\midrule
acq\_func & EI & EI & PI & PI\\
init\_design & Random & LHD & Sobol & Sobol\\
n\_configs\_x\_params & 1 & 1 & 4 & 2\\
y\_trans & y & y & LogScaled(y) & y\\
lcb\_par & - & - & - & - \\ 
par & 0.0 & 0.08 & 1.00 & 0.51 \\
scale\_inv\_perc & - & - & - & -\\
scale\_log\_perc & - & - & 20 & -\\
rand\_prob & 0.0 & 0.06 & 0.07 & 0.10 \\
\midrule
ard & False & True & True & True\\
kernel & rbf & matern & matern & rbf\\
ls\_lower\_bound & 0.0025 & 0.0005 & 0.0042 & 0.0298 \\
ls\_upper\_bound & 7.39 & 3.76 & 4.57 & 72781.34 \\
tune\_noise & False & True & True & False \\
\bottomrule
    \end{tabular}
    \label{tab:my_label}
\end{table}

\clearpage
\begin{table}[h]
    \centering
    \caption{Best found configuration for target-BO using GP-MAP. Values were rounded to the second or fourth digit for increased readability.}
    \begin{tabular}{lr r r r}
\toprule
 & {Default}  & artificial & HPO SVM & ParamNet \\
\midrule
acq\_func & EI & EI & LCB & PI\\
init\_design & Random & Random & LHD & Sobol\\
n\_configs\_x\_params & 1 & 9 & 3 & 1\\
y\_trans & y & LogScaled(y) & LogScaled(y) & LogScaled(y)\\
lcb\_par & - & - & 0.17 & -\\
par & 0.0 & 0.03 & - & 0.42 \\
scale\_inv\_perc & - & - & - & -\\
scale\_log\_perc & - & 11 & 9 & 2\\
rand\_prob & 0.0 & 0.01 & 0.02 & 0.02 \\
\midrule
ard & False & False & True & True\\
kernel & rbf & matern & rbf & rbf\\
ls\_lower\_bound & 0.0025 & 0.0030 & 0.0062 & 0.0001 \\
ls\_upper\_bound & 7.39 & 152.42 & 1.78 & 34027.35 \\
tune\_noise & False & False & False & True\\
\midrule
prior\_cov\_ampl & Lognormal & Gamma & Lognormal & Lognormal\\
prior\_ls & None & Gamma & None & Gamma\\
prior\_noise & Horseshoe & Gamma & Horseshoe & Gamma\\
prior\_cov\_ampl\_gamma\_a & - & 0.14 & - & -\\
prior\_cov\_ampl\_gamma\_scale & - & 0.22 & - & -\\
prior\_cov\_ampl\_lognormal\_sigma & 1.0 & - & 0.01 & 0.05 \\
prior\_ls\_gamma\_a & - & 3.05 & - & 1.35 \\
prior\_ls\_gamma\_scale & - & 0.03 & - & 0.57 \\
prior\_noise\_gamma\_a & - & 0.81 & - & 0.15 \\
prior\_noise\_gamma\_scale & - & 0.48 & - & 0.68 \\
prior\_noise\_horseshoe\_scale & 0.1 & - & 0.05 & -\\
\bottomrule
    \end{tabular}
    \label{tab:my_label}
\end{table}

\begin{table}[h]
    \centering
    \caption{Best found configuration for target-BO using GP-MCMC. Values were rounded to the second or fourth digit for increased readability.}
    \begin{tabular}{lr r r r}
\toprule
 & {Default}  & artificial & HPO SVM & ParamNet \\
\midrule
acq\_func & EI & LCB & PI & EI\\
init\_design & Random & Random & LHD & Random\\
n\_configs\_x\_params & 1 & 4 & 4 & 6\\
y\_trans & y & y & LogScaled(y) & InvScaled(y)\\
lcb\_par & - & 81.28 & - & -\\
par & 0.0 & - & 0.48 & 0.08 \\
scale\_inv\_perc & - & - & - & 6\\
scale\_log\_perc & - & - & 1 & -\\
rand\_prob & 0.0 & 0.04 & 0.04 & 0.03 \\
\midrule
ard & False & True & False & True\\
kernel & rbf & matern & rbf & matern\\
ls\_lower\_bound & 0.0025 & 0.0013 & $<$0.0001 & $<$0.0001 \\
ls\_upper\_bound & 7.39 & 3.03 & 38.04 & 75097.66\\
tune\_noise & False & True & False & True\\
\midrule
prior\_cov\_ampl & Lognormal & Lognormal & Gamma & Gamma \\
prior\_ls & None & None & None & None\\
prior\_noise & Horseshoe & Horseshoe & Horseshoe & Gamma \\
prior\_cov\_ampl\_gamma\_a & - & - & 0.46 & 2.17 \\
prior\_cov\_ampl\_gamma\_scale & - & - & 0.39 & 0.93 \\
prior\_cov\_ampl\_lognormal\_sigma & 1.0 & 0.72 & - & -\\
prior\_ls\_gamma\_a & - & - & - & - \\
prior\_ls\_gamma\_scale & - & - & - & - \\
prior\_noise\_gamma\_a & - & - & - & 1.26 \\
prior\_noise\_gamma\_scale & - & - & - & 0.01 \\
prior\_noise\_horseshoe\_scale & 0.1 & 0.39 & 0.06 & -\\
\bottomrule
    \end{tabular}
    \label{tab:my_label}
\end{table}

\newpage
\bibliographystyle{abbrvnat}
\bibliography{strings,lib,local,proc}